# Polynomial Value Iteration Algorithms for Deterministic MDPs

**Omid Madani**
Department of Computing Science
University of Alberta
Edmonton, AB
Canada T6G 2E8
madani@cs.ualberta.ca

## Abstract

Value iteration is a commonly used and empirically competitive method in solving many Markov decision process problems. However, it is known that value iteration has only pseudo-polynomial complexity in general. We establish a somewhat surprising polynomial bound for value iteration on deterministic Markov decision (DMDP) problems. We show that the basic value iteration procedure converges to the highest average reward cycle on a DMDP problem in $\theta(n^2)$ iterations, or $\theta(mn^2)$ total time, where $n$ denotes the number of states, and $m$ the number of edges. We give two extensions of value iteration that solve the DMDP in $\theta(mn)$ time. We explore the analysis of policy iteration algorithms and report on an empirical study of value iteration showing that its convergence is much faster on random sparse graphs.

## 1 Introduction

Markov decision processes offer a clean and rich framework for problems of control and decision making under uncertainty[BDH99, RN95]. Infinite-horizon fully observable MDP problems (MDPs) are classic optimization problems in this framework. Not only are the MDP problems significant on their own, but solutions to these problems are used repeatedly in solving problem variants such as stochastic games, and partially observable MDPs [Sha53, Han98]. Preferred methods for solving MDP problems use dynamic programming strategies, and in particular often contain a so-called value iteration or policy iteration loop [Put94, Lit96, Han98, GKP01]. These methods converge to optimal solutions quickly in practice, but we know little about their asymptotic complexity. It is known, however, that algorithms based on value iteration have no better than a pseudo-polynomial[1] run time on MDP problems [Tse90, Lit96].

In this paper, we analyze the basic value iteration procedure on the deterministic MDP problem under the average reward criterion, or the DMDP problem, and we establish several positive results. The DMDP problem is also known as the maximum (or minimum) mean cycle problem in a directed weighted graph [AMO93]. In solving DMDPs, we are often interested in finding a highest average weight cycle (an *optimal cycle*), or the average weight of such a cycle (the *highest mean*). A policy in this problem is simply a subgraph in which each vertex (state) has a single out-edge (action choice) leading by a directed path to an optimal cycle. Just as is the case for general MDPs, the DMDP has both direct and indirect applications, for example in solving network flow problems and in system-performance analysis [AMO93, DG98, CTCG+98].

We establish that in graphs with $n$ vertices and $m$ edges, basic value iteration converges to an optimal cycle in a DMDP in $O(n^2)$ iterations, *irrespective* of the initial assignment of values. This is somewhat unexpected considering that value iteration is generally pseudo-polynomial. We also show that the bound is tight by giving an example on which value iteration takes $\Omega(n^2)$ iterations. We note that while an optimal cycle is found in polynomial time, examples exist where value iteration still takes pseudo-polynomial time to converge to an optimal *policy*. This occurs because it may take many iterations before states that do not reside on an optimal cycle choose their optimal action (Section 3.1). Therefore, in a sense, value iteration is 'almost' polynomial in finding optimal policies for DMDPs. Nevertheless, finding the optimal cycles is the main task in DMDPs, and we expect that value iteration converges much faster in practice. Our experiments on random graphs show that value iteration converges exponentially faster than the $n^2$ worst-case bound would suggest (see Section 6).

The insight from the analyses allows us to show that, by making small modifications to value iteration, optimal cycles (and policies) can be found in $\theta(n)$ iterations. As each iteration takes $\theta(m)$ time, this gives an algorithm that ties the only other algorithm with the same run-time of $\theta(mn)$ [Kar78, AMO93]. An algorithm described here has an additional interesting property that it is distributed: each vertex performs a simple local computation and need only communicate to its immediate neighbors via its edges, and by $\theta(n)$ iterations, all vertices will know the highest mean. With $\theta(n)$ more iterations, the optimal cycle would also form. We remark that the algorithmic technique developed here also extends to give polynomial algorithms for more general problem classes where edges may have two parameters: a probability or (time) cost in addition to a reward [Mad02a].

---

[1] An algorithm has pseudo-polynomial run time complexity, if it runs in time polynomial in the unary representation of the numbers in the input, but exponential in the binary representation.

We also give a polynomial algorithm that is similar to the multi-chain policy iteration algorithm for the DMDP [Put94]. The polynomial bound proof for this algorithm is identical to that for simple value iteration, but we conjecture that the bound is not tight. We describe the similarity to policy iteration and the difficulty of analyzing policy iteration on DMDPs, and indicate promising ways of addressing the open problems.

We investigate the convergence of value iteration on random sparse graphs. The experiments suggest that value iteration converges to optimal cycles in an expected $O(\log n)$ many iterations, i.e., exponentially faster than what the worst case bound indicates. These experiments provide valuable insights into why such algorithms have excellent performance in practice.

We begin the paper with problem definitions and notation in the next section. In Section 3, we describe our analysis of value iteration. Section 4 presents two modified algorithms with $\theta(mn)$ time. Next, we give our results on policy iteration, and, in Section 6, report on our experiments and discuss previous empirical studies of algorithms on DMDPs. Section 7 concludes with a summary and a discussion of open problems and future directions. Throughout the paper, we have tried to describe the line of argument and have given proofs sketches for the important steps. Complete proofs with more explanations and an expanded empirical section appear in [Mad02b].

## 2 Preliminaries

We give the graph theoretic definition of the DMDP problem here to save space. Let $G = (V, E, r)$ be a directed graph with $n$ vertices and $m$ edges, where $r$ is a function from the edge set $E$ into real numbers such that for an edge $e$ in $E$, $r(e)$ is the reward of $e$. Each vertex has at least one single out-edge. A *walk* $w = e_1, e_2, \cdots, e_k, k \geq 1$ is a progression of edges such that the end vertex of $e_i$ is the start vertex of $e_{i+1}$. A walk may have repeated edges. We call an edge $(u,v)$ (resp. walk) connecting vertex $u$ (start vertex $u$) to end-vertex $v$ a $u$-$v$ edge (resp. walk). For a walk $w$, let $|w|$ denote the number of edges in $w$, let $R(w) = \sum_{1 \leq i \leq k} r(e_i)$ be its *total reward*, and let $\bar{R}(w) = \frac{R(w)}{k}$ be its *average reward* or *mean*. A *cycle* is a walk where the start vertex of $e_1$ is the same vertex as the end of $e_k$, and no other vertex is repeated. Let $\mu^* = \max_c \bar{R}(c)$, where $c$ ranges over all cycles in $G$. We call $\mu^*$ the *maximum mean* and in solving a DMDP problem we are interested in either the problem of computing $\mu^*$ or finding an *optimal cycle* with mean $\mu^*$ (the problems are equivalent).

The DMDP problem was shown solvable in $O(mn)$ time by Karp [Kar78]. Later, algorithms were given with run times of $O(mn \log n)$, and these algorithms are believed to be faster than (unmodified) Karp's algorithm in practice [YTO91], as Karp's algorithm takes $\theta(mn)$ irrespective of underlying graph. To the best of our knowledge, all the algorithms with known $O(mn)$ prior to our work [HO93, DG98], are modifications of Karp's algorithm.

The basic value iteration process is shown in Fig. 1. Let $t = 0, 1, 2, \cdots$ denote time points, where time $t$ is immediately before iteration $t+1$ of the algorithm. Let $x_v^{(t)}$ denote the *value* of vertex $v$ at time $t$, thus $x_v^{(0)}$ is its initial value. Then the *value* of a $u$-$v$ edge $e$ in iteration $t, t \geq 1$ is $r(e) + x_v^{(t-1)}$. Each vertex $u$ at each iteration $t$ performs the following computation to obtain its new value $x_u^{(t)}$:

$$x_u^{(t)} \leftarrow \max_{u\text{-}v \text{ edge } e} r(e) + x_v^{(t-1)}, \quad t \geq 1. \tag{1}$$

We call a subgraph of $G$ with out-degree exactly one for each vertex a *policy*. Thus the choice of an out-edge for each vertex in each iteration of value iteration defines a policy which we say value iteration *visits*[2]. We say a vertex changes edges or simply *switches* at an iteration if the choice of edge according to eq. 1 changes from the previous iteration. The following assumption is important for the correctness of the algorithms: for any vertex, if the chosen edge from previous iteration ties in the best value, then that edge is chosen again. In other words, value iteration is 'lazy' in changing the policy from one iteration to another. Ties in the first iteration or when the previous edge is not a highest valued edge may be broken arbitrarily. Note that each iteration of value iteration takes $O(m)$ time. Fig. 2 shows the first two iterations of value iteration where the vertices are assigned zero values initially. The sequence of values for vertex $u$ in the figure is: $\langle x_u^{(t)} \rangle = \langle x_u^{(0)}, x_u^{(1)}, x_u^{(2)}, \ldots \rangle = \langle 0, 5, 11, 14, \cdots \rangle$.

---

[2]Interestingly, $\mu^*$ has a related characterization of being the maximum eigenvalue of the $n \times n$ matrix $A$ of edge weights under max-plus operations: multiplication and summation in $Ax$ translate to summation and taking the maximum respectively [CTCG+98]. This in turn corresponds to the dynamic programming operation in eq. 1. Our result shows that for any value vector $x$, the vector $y = A^{n^2} x$ is special: performing value iteration on $y$ for at most $n$ iterations visits a policy containing an optimal cycle.

1. Each vertex begins with a value
2. Repeat
3. Each vertex chooses its highest valued out-edge e and obtains the value of e

Figure 1: Pseudocode for value iteration.

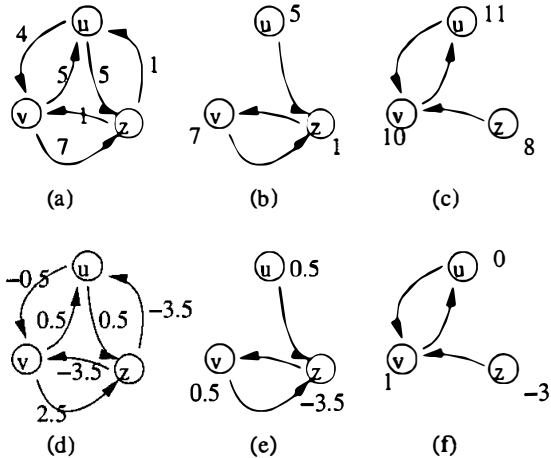

Figure 2: (a) A three vertex DMDP graph shown with edge rewards. (b) and (c) First two iteration of value iterations where vertices start with zero initial values, and visited policies and vertex values after each iteration are shown. (d), (e), and (f) The corresponding mean-zero parallel graph (Sect. 3), and the first two value iterations.

## 3 Polynomial Convergence of Value Iteration

Here, we first show that analyzing value iteration on a transformed DMDP graph where $\mu^* = 0$, which we refer to as a *mean-zero* graph, suffices in showing polynomial convergence, then we analyze value iteration on mean-zero graphs.

Consider value iteration as it proceeds on a graph $G = (V, E, r)$, where vertices are initialized with arbitrary values. The *value* of a $u$-$v$ walk $w$ at an iteration $t \geq |w|$ is the sum of its total reward and value of end-vertex $v$ $|w|$ time points ago: $R(w) + x_v^{(t-|w|)}$. For example, in Fig. 2, the value of $u$-$v$ walk composed of a single $u$-$v$ edge in iteration one is 4 and in iteration two is 11, which is also the value of the $u$-$z$ walk $w = (u,v), (v,z)$ in iteration two. The next lemma relates the value of walks and the value of vertices. It can be shown by induction on time $t$ (or length of walks).

**Lemma 3.1** $x_v^{(t)}$ *is the maximum value over the values of all walks of length $t$ with start vertex $v$.*

The *history walk* of a vertex $v$ at a given iteration is, informally, the walk formed following the sequence of edge selections of value iteration starting at $v$ and going back in time. More precisely, at iteration $t$, if vertex $v$ has edge $e$ to $u$, then its history walk of one time step consists of edge $e$. Its history walk of $k > 1$ steps is $e$ concatenated with history walk of length $k - 1$ steps for vertex $u$ at iteration $t - 1$. The history walk is necessarily a maximum valued walk by Lemma 3.1. In Fig. 2, the history walk of length two for vertex $z$ in iteration 2 is $(z, v), (v, z)$ and for vertex $v$ it is $(v, u), (u, z)$.

Let us call two graphs as *parallels* of one another if they have identical vertex and edge sets, but the reward function $r$ in one graph is a constant offset of the reward function $r'$ for the other, or $\forall e \in E, r'(e) = r(e) - \mu$, for some constant $\mu$. The average reward of a cycle is offset by $\mu$ as well, and therefore cycles keep their relative merits in the corresponding graphs, and in particular optimal cycles remain the same under such a transformation of edge rewards[3]. Furthermore, value iteration behaves identically on parallel graphs, meaning that each vertex selects the same edge in either problem at any iteration (subject to ties), as a consequence of the next lemma:

**Lemma 3.2** *Consider value iteration started with equal initial values on a pair of parallel graphs $G$ and $G'$, with reward functions $r$ and $r'$ respectively where $r'(e) = r(e) - \mu$, for some constant $\mu$. Then if $x_v^{(t)}$ is the value of vertex $v$ at time $t$ in graph $G$, $x_v^{(t)} - \mu t$ is the value of vertex $v$ in $G'$ at time $t$.*

**Proof.**(sketch) The two values of any walk of length $t$, in particular a history walk, in two parallel graphs are different by $\mu t$. We can then use Lemma 3.1. □

Now, given a graph with maximum mean $\mu^*$, the parallel graph with offset $\mu^*$ has maximum mean 0 (Fig. 2d). As a consequence of (1) the optimal cycles being identical in parallel graphs, and (2) the identical behavior of value iteration on parallel graphs, the properties on the structure of history walks and policies that we show hold on the mean-zero graph in the next section (*e.g.* lemma 3.6) also hold in the general case. Fig.3 gives a high level picture of the identical behavior of value iteration on parallel graphs and rate of convergence to optimal cycles and policies.

### 3.1 Value Iteration on Mean-Zero Graphs

The line of argument in this section is roughly as follows. We show that for any vertex, its sequences of values has a maximum and the maximum is reached in no more than $n$ iterations (Lemmas 3.3 and 3.4). For vertices in the mean-zero (optimal) cycle, once they obtain the maximum values (and a collection of other upper-bound values to be defined), they can "keep" these values by choosing the edges of the optimal cycle, and we show that this is in fact what happens (Lemmas 3.5, 3.6, and 3.7).

---

[3]The same transformation is used in [Kar78].

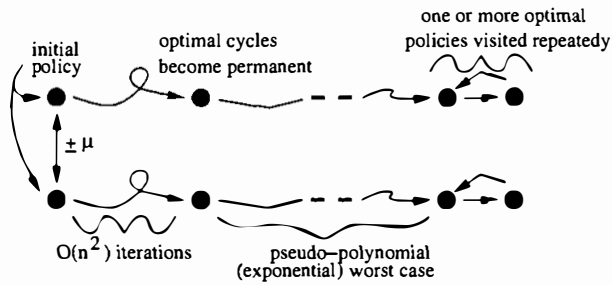

Figure 3: Value iteration visits the same sequence of policies when it starts with identical initial values on parallel graphs. Some policies (but not vertex values) may repeat on the way to optimal policies. Convergence can take exponentially many iterations, but only $O(n^2)$ iterations for the formation of optimal cycles.

In the mean-zero case, no cycle has a positive reward, from which the following lemma follows.

**Lemma 3.3** *In the mean-zero case, if a vertex $v$ has a history walk of length $j$ to itself at an iteration $t$, then $x_v^{(t)} \leq x_v^{(t-j)}$, and $x_v^{(t)} < x_v^{(t-j)}$ if the cycle is suboptimal, i.e., has negative mean.*

Now, consider the sequence of values of a vertex $v$, $\langle x_v^{(t)} \rangle = \langle x_v^{(0)}, x_v^{(1)}, \ldots \rangle$, as value iteration proceeds on a mean-zero graph. The next lemma, which is central to our results, bounds the latest time an increase in the maximum value can occur over the sequence $\langle x_v^{(t)} \rangle$ and also its subsequences. Take $p$ in the lemma as the period or the interval at which we look at the values in the sequence $\langle x_v^{(t)} \rangle$. Consider the simplest $p = 1$ case, i.e., look at all the values of the sequence). The lemma states that for any vertex, an increase in its maximum value–over its values seen so far–can occur only in the first $n$ iterations. In other words, (1) such a sequence of values has a maximum and (2) it first occurs in the first $n$ iterations. Similarly, with $p = 2$, we consider the even and odd subsequences. The lemma states that the two highest values, one over the even subsequence ($\langle x_v^{(2t)} \rangle$) and the other over the odd subsequence ($\langle x_v^{(2t+1)} \rangle$), appear in at most $2n$ iterations. Note that one of the two highest values is the highest value over any subsequence ($p = 1$) and must appear in the first $n$ iterations. Similar results hold for higher periods $p$.

**Lemma 3.4** *For any $p \geq 1$, at any time $l \geq p$, if the following ("dominance") property holds:*

$$\forall i < l, \text{ such that } i \equiv l \pmod{p}, x_v^{(i)} < x_v^{(l)}, \qquad (2)$$

*then $l \leq pn$.*

**Proof.** (sketch) Assume the dominance property holds for vertex $v$ at time $l$. Consider the subsequence $w'$ of vertices formed by examining the history walk $w$ of $v$ after every $p$ steps (for example, if $p = 2$ and the vertex sequence in history walk $w$ is $\langle v, z, u, v, y, \cdots \rangle$ then vertices in $w'$ are $\langle v, u, y, \cdots \rangle$). Therefore, $l = |w| = p|w'|$. We can see that $w'$ cannot repeat a vertex due to the dominance property and Lemma 3.3. It follows that $|w'| \leq n$, or $l \leq pn$. □

Lemma 3.4 has many consequences. For any integer $k \geq 1$ and any vertex $v$, let $x_{v,k}[j], 0 \leq j < k$, denote the highest value a vertex $v$ obtains in any iteration $t \equiv j \mod k$. These $k$ values are well-defined (bounded) by Lemma 3.4. We call these $k$ values the *highest $k$ values* of $v$. For example, for the mean-zero parallel graph in Fig. 2d, vertex $v$ obtains the following value sequence $\langle 0, 0.5, 1.0, 0.5, 1.0, \cdots \rangle$ (with zero initial vertex value assignment), and its two highest values are $x_{v,2}[0] = 1.0$, and $x_{v,2}[1] = 0.5$. Assume vertex $v$ is in a mean-zero (optimal) cycle $c$. Since $c$ has mean zero, it follows from Lemma 3.1 that once $v$ obtains a highest value from its $|c|$ highest values, it gets it back every $|c|$ iterations. Note that some of the highest values may be equal. Let $x_v^*$ be the highest value $v$ obtains ever. As another consequence of Lemma 3.4, $v$ obtains value $x_v^*$ in the first $n$ iterations and its highest $k$ values in no more than $kn$ iterations. We show basically that once the $|c|$ highest values reach the vertices of a mean-zero cycle $c$, the vertices of the cycle do not need to "deviate" from that cycle, i.e., they can always choose the edges of the cycle in every subsequent value iteration (Lemma 3.7).

The next two lemmas help us establish convergence. We observe that once a vertex $v$ in $c$ obtains its highest value $x_v^*$, its immediate neighbor in $c$ must obtain its highest value in the next iteration, and in general some vertex in $c$ obtains its highest value in every subsequent iteration. Lemma 3.5 is the generalization of this property to all $|c|$ highest values as defined above. The first statement of Lemma 3.6 is a consequence of Lemma 3.3. Then the second statement follows using in addition Lemmas 3.5 and 3.4.

**Lemma 3.5** *Assume vertex $u$ has an edge to $v$ in a mean-zero (i.e. optimal) cycle $c$. Then whenever $v$ obtains its $j$th highest value $x_{v,|c|}[j]$, $u$ obtains its $j + 1$st highest value $x_{v,|c|}[j + 1]$ in the following iteration.*

**Lemma 3.6** *The history walk of a vertex $v$ in an optimal cycle $c$, whenever $v$ obtains its highest value, includes only optimal cycles. At any iteration $t \geq n$, the history walk of some vertex on an optimal cycle includes only optimal cycles.*

**Lemma 3.7** *When all vertices in all optimal cycles have obtained their highest values, after at most $n$ more iterations, some optimal cycle appears in all subsequent visited policies.*

**Proof.** When vertices of an optimal cycle find all their highest values, if some vertex chooses the out-edge in the cycle, it will not switch (change edge) again due to our assumption of "lazy" policy change. When all vertices of optimal cycles find their highest values, whenever a vertex $v$ obtains $x_v^*$, its history walk can only contain mean-zero (in general optimal) cycles by 3.6. Consider the first such cycle in the walk. As its vertices have found their highest values, those vertices will not switch again. □

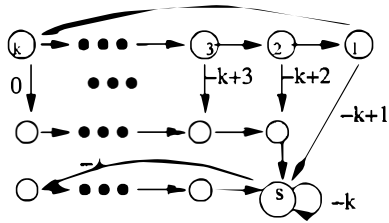

Figure 4: An example graph where it takes $\theta(n^2)$ iterations for the optimal cycle to form for the first time. All edges without a displayed reward have zero reward. The numbered vertices in top row form the optimal cycle which has $k$ vertices and zero mean. The remaining two rows have $k - 1$ vertices each. Vertex $s$ is initialized with zero and all others are initialized with $-k^3$.

If there are multiple optimal cycles, examples exist where some vertices may repeatedly change edges forever, but it can be shown that eventually all vertices will have some path to some optimal cycle in any visited policy. However, consider a vertex with a high reward edge to a cycle of mean -1, and another edge to a mean zero (optimal) cycle. It is not hard to see that eventually it will choose the edge to the mean-zero cycle, but this could take many iterations. This example can be formalized to show that the worst-case number of iterations to optimal policies is pseudo-polynomial (see for example [ZP96]).

It is also not hard to give an example for which the time until all the highest values arrive at the vertices of an optimal cycle can be $\theta(n^2)$. This would mean the time until an optimal cycle becomes fixed is $\theta(n^2)$ in the worst case. Fig. 4 shows that even the *first time* formation of an optimal cycle takes $\theta(n^2)$ time, in the worst case. In the given graph, vertex $s$ is initialized with zero, and all other vertices may be assigned any value less than $-k^3$. This example is explained further in the expanded paper [Mad02b].

As a consequence of Lemmas 3.7, 3.4, 3.2, and the example graph given, we obtain the following theorem on value iteration on DMDPs.

**Theorem 3.8** *Value iteration converges to an optimal cycle in a DMDP problem in $\theta(n^2)$ iterations.*

We remark that a common variation of value iteration, referred to as Gauss-Siedel value iteration [Put94], does not necessarily converge to an optimal cycle. In this variation, the vertices are numbered, the vertex values are updated in order in each iteration, and the new value of a vertex is used as soon as it becomes available. Note that this is a natural implementation of value iteration on a sequential machine. But history walks in this case can be longer than the number of iterations, and Lemma 3.2 in particular breaks for Gauss-Siedel value iteration, i.e. Gauss-Siedel value iteration does not have identical behavior on parallel graphs. However, it can be shown that it converges to some cycle, and moreover the properties of value iteration on mean-zero graphs still hold for Gauss-Siedel.

## 4 Algorithms Based on Histories

We can still compute the optimal cycle using basic value iteration in $O(n)$ iterations even though convergence takes $\theta(n^2)$ iterations. Lemma 3.6 shows us how. If we keep track of the edge chosen by each vertex in each iteration for the first $n$ iterations as value iterations progresses, we can reconstruct the cycles in the history walks, and some cycle must be optimal by Lemma 3.6. Searching for the optimal cycle takes $n^2$ time, thus the algorithm takes $O(n^2 + mn) = O(mn)$ time, but unfortunately requires $\theta(n^2)$ space.

We next describe a variation that reduces the space back to linear. The algorithm has the desirable property that just like value iteration it has a distributed nature: each vertex performs a simple local computation until all vertices discover the optimal mean. This algorithm also works in two phases, the first phase being simple value iteration for $n$ iterations. The second phase takes $n$ iterations as well, but each vertex performs an additional computation in addition to updating its value and edge choice. In each iteration after $n$, each vertex keeps track of not only its current value and chosen edge, but updates parameters characterizing its *super edge* as well. Super edges summarize history walks, and may be viewed as *packets* sent along edges. Each super edge is either 'dropped', or is updated and passed along in each iteration. When a vertex discovers that a super edge was sent by itself, it computes the average value of the cycle corresponding to the super edge from the parameters of the super edge, and updates the current highest mean $\mu$ found so far. The highest mean found in the second phase is the optimal $\mu^*$.

A super edge has three parameters $(v, l, r_s)$, where $v$ is the vertex it ends in, $l$ is the number of edges in the super edge, and $r_s$ is its total reward. At any iteration, such as the beginning of the second phase, the super edge of a vertex may be undefined. Vertex $u$ computes its super edge at iteration $t$ as follows. Assume $u$ chooses the $u$-$v$ edge $e$ in iteration $t$. In case $u \neq v$, if the super edge for $v$ of iteration $t - 1$ is undefined, the super edge for $u$ is defined to be $(v, 1, r(e))$. If $v$ has super edge $(z, l, r_s)$, and $u \neq z$, the super edge for $u$ is $(z, l + 1, r(e) + r_s)$. Otherwise, when $u = v$ or $u = z$, vertex $u$ has obtained a *cyclic* super edge, and the mean is respectively $r(e)$ or $\frac{r(e)+r_s}{l+1}$. In this case, the running estimate $\mu$ is updated if necessary, and vertex $u$ marks its current super edge undefined. As an example, if ver-tices where to begin keeping track of super-edges starting from iteration 1 in Fig. 2a, then the super-edges for vertex $u$ at iterations 1 and 2 would be respectively $(z, 1, 5)$ and $(z, 2, 11)$. If vertices began keeping track of super-edges starting from iteration two, the super-edges of vertex $u$ at iterations 2 and 3 would be $(v, 1, 4)$ and $(u, 2, 9)$ respectively, and at end of iteration 3, vertex $u$ would update the highest mean found so far if necessary (in this case 9/2), and mark its current super-edge as undefined.

The algorithm takes $2n$ iterations and each iteration takes constant time per edge, thus the run time is $O(mn)$, with only $O(n)$ extra space. Correspondence made between super edges and history walks, and Lemma 3.6 establish the correctness. A subtlety is when there are multiple optimal cycles and ties in edge selections occur. In this case we assume a vertex chooses the edge whose end-vertex has a super edge with lowest numbered vertex. In the beginning of the second iteration where no vertex has a super edge, we assume ties are broken based on the lower numbered end-vertex. We call this rule the "lowest-index" rule.

The following lemmas establish the properties of super edges and lead to correctness of the algorithm:

**Lemma 4.1** *At any time point, the super edge of length $l$ for a vertex $v$, if any, corresponds to the history walk of length $l$ for vertex $v$ at that time.*

Let $\mu$ be the highest mean over the averages of cyclic super edges discovered in the second phase. Lemma 4.2 can be shown by noting that a walk with the same start and end vertex, possibly with two or more cycles inside, where each has mean no greater than $\mu^*$, does not have a mean greater than $\mu^*$.

**Lemma 4.2** *No cyclic super edge has average reward greater than the optimal mean value $\mu^*$. Therefore $\mu \leq \mu^*$ throughout the algorithm.*

That some cyclic super edge computed in the second phase has mean $\mu^*$, is not hard to see when the optimal cycle is unique, as the highest value in the mean-zero graph is created and traces the optimal cycle after the first $n$ iterations. In case of multiple optimal cycles, ties in edges may not be broken arbitrarily, otherwise examples show that no cyclic super edge corresponding to an optimal cycle is created in the second $n$ iterations. But the lowest-index rule prevents this. We expect other easier rules–for example if each vertex breaks ties consistently locally–also give correct algorithms.

**Lemma 4.3** *Assume the lowest-index rule is used in breaking ties. Then some vertex obtains a cyclic super edge corresponding to an optimal cycle in the second $n$ iterations.*

Correctness of the algorithm, which we shall refer to as the *history-walk* algorithm, follows.

---

1. Begin with an arbitrary policy
2. Repeat until no new cycle is discovered
3.     Update edge rewards, edge choices and vertex values
4.     Apply value iteration until a new cycle is discovered or until $n$ iterations.

Figure 5: Generic phased policy iteration.

**Theorem 4.4** *The history-walk algorithm takes $\theta(mn)$ time and uses $\theta(n)$ space in finding the optimal mean $\mu^*$.*

A natural question is whether vertices may begin keeping track of super edges before iteration $n$. The answer is negative in the worst case, at least as far as the algorithm just described, as cyclic super edges may not ever form in this case [Mad02b].

## 5 On Policy Iteration Algorithms

Consider the following change to value iteration, which we call *augmented value iteration*: At each iteration, the cycles in the visited policy are identified and the highest cycle mean $\mu$ is computed. Then each vertex $v$ gets a self-arc ( $v$-$v$ edge) with the same reward (or mean) $\mu$ so that $v$ can choose the self-arc in subsequent iterations. That this algorithm finds the optimal mean $\mu^*$ in at most $O(n^2)$ iterations from the same convergence arguments used for value iteration, but the bound may not be tight. On the other hand, this algorithm is very similar to the so-called multi-chain policy iteration algorithm for average reward MDP problems [Put94, CTCG+98]. These algorithms can be viewed as working in *phases*, as shown in Fig. 5, where each phase begins with using the mean of the recently discovered cycle, and updates edge rewards and vertex values appropriately, and then begins a series of value iterations until another cycle is found. The algorithms differ on how they update values, edge choices, and edge rewards, but they all guarantee that the next cycle discovered will have higher mean than the last. In the augmented value iteration algorithm, vertex values are not changed, however, new self-arcs with most recently found cycle mean $\mu$ are added. Alternatively, in augmented value iteration, we may subtract $\mu$ from all edge rewards, but keep zero reward self-arcs for each vertex: this does not change the optimal cycle, nor the behavior of value iteration by Lemma 3.2.

In policy iteration, in addition to subtracting $\mu$ from edge rewards, each vertex redirects itself to the cycle with mean $\mu$, that is, the algorithm finds a policy so that all ver-tices have a path[4] to cycle $c$. Vertices are then reassigned values as follows: an arbitrary vertex $v$ in the cycle of the current

---

[4]Without loss of generality we may assume the graph is strongly connected. Otherwise, the algorithm performs this for each component.

policy is assigned 0, and all others get the total re-ward of their path to $v$.

The reassignment of values to vertices in policy iteration appears to make analysis difficult. However, in work in progress we have shown that variants of augmented value iteration where vertices get reassigned zero values (or any value vector that remains constant across phases) before value iteration begins in each phase, terminate within a polynomial number of phases and therefore run in polynomial time. Just as in policy iteration, in these algorithms vertex values can only increase during value iteration within a phase, and the cycles discovered improve from one phase to the next. The basic behavior seen from the results is that vertices behave almost identically, in terms of the edges they choose in corresponding iterations from one phase to the next. The exception is that progressively more vertices have zero increase in value and thus stop switching edge choices, with each subsequent phase. These results, however, make use of the fact that vertices begin with the same value vector (for example zero) in each phase, which simplifies comparison between phases and aids analysis. Relaxing such constraints, and improving the bounds for these algorithms may provide fruitful insights on the path to establishing efficiency of policy iteration algorithms.

## 6 Behavior on Random Graphs

The algorithms we have developed require at least a linear number of iterations. However we suspected that on random graphs, relatively few iterations of value iteration would suffice in finding the maximum mean. We explored these questions on random graphs where every vertex has two out-edges, the end-vertex of each edge is chosen uniformly at random from the remaining $n-1$ vertices, and the reward of edges were chosen uniformly at random in $[0, 1]$. We tested on sparse graphs only, as the number of actions (edges) per vertex is usually small in MDP and many other problems– $m$ is often a linear function of $n$ –which leads to problems on sparse graphs. The averages for graphs of size $n$ were obtained over samples of size maximum of 500 and $n$. Fig. 6a suggests that both the growth of the expected optimal cycle length and the number of iterations until an optimal cycle is first formed in a policy increase as polylogarithmic functions of graph size[5] (apparently, bounded by $O(\log n)$ and $O(\log^2 n)$ respectively).

One way to compute the optimal mean quickly is to test the current visited policy periodically and compute whether average reward of the cycle(s) in the policy is optimal. An efficient way to test this is to subtract the candidate mean from all edge rewards, and use an efficient implementation of the Bellman-Ford shortest paths algorithm to detect the presence of positive cycles [AMO93, CLR92]. If there are no positive cycles, the candidate mean is the maximum. While the shortest path detection algorithm also has $O(mn)$ run time, empirically it is very efficient and may have linear expected time [KB81]. Fig. 6b verifies our expectation. It shows the run times for two algorithms that test periodically after $\log n$ initial value iterations. These tests were performed on a Pentium III/500 with 128 megabytes of RAM with small load. One algorithm uses super edges (find-in-history) and another simply checks the cycles formed in the policies (find-in-policy). The run times of both algorithms are close to linear time as expected. The plots show that the find-in-policy version seems to perform better, probably due to its lower overhead and that find-in-history also has to wait a number of iterations until an optimal cycle is formed in a super edge.

---
[5]We expect that the relatively high average length of optimal cycles for size 25 graphs is due to the small size of the graphs. With larger graph sizes the limiting distribution of the random variable seems to kick in.

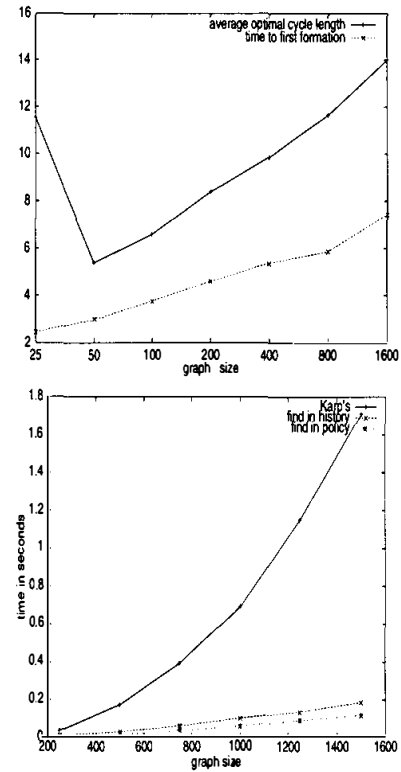

Figure 6: (a) Averages of optimal cycle lengths and the first iteration until an optimal cycle is found. (b) A comparison of the run times.

Many DMDP algorithms including policy iteration, but not value iteration, are tested on several graph families in [DG98], and they conclude that policy iteration is the fastest. We expect that the algorithms given here will be very competitive empirically as well due to their low overhead. In particular, the augmented value iteration algorithm (Sec. 5) may have a lower overhead than policy iteration as it need not compute path values for vertices in each phase: it simply continues from the current vertex values. In future work, we will further compare the performance of these algorithms on DMDPs and related problems.

## 7 Discussion

We noted that value iteration does not solve the problem of finding an optimal policy in polynomial time. Thus the DMDP problem may be considered a borderline problem on which value iteration is almost polynomial. On shortest path problems with no negative cycles, and on DMDPs where all cycles share a single vertex, value iteration finds an optimal policy in polynomial time. Higher up in the problem hierarchy, on general stochastic MDP problems and several subclasses where the degree of stochasticity is limited, it takes exponential time to converge to optimal cycles or policies. Perhaps the closest problem to the DMDP is the discounted deterministic MDP problem. On these problems, as the discount $\beta$ approaches 1 an optimal cycle becomes the same as the highest average reward cycle [Put94], and with small $\beta$ the problem is easy to approximate. Therefore an approximation property such as the following may hold: $O(n^2)$ runs of value iteration starting with any initial vector is sufficient to converge to approximately optimal cycles in discounted deterministic MDPs.

This work was motivated by the analysis of policy iteration on MDPs and is in line with developing a complete picture on the efficiency of value and policy iteration algorithms on MDPs. Studying simpler problem classes can lead to techniques of algorithm design and analysis applicable to more general problems, as well as giving a better understanding of where new ideas are needed when such techniques fail to generalize. Our hope is that the gaps in our understanding of whether and why these algorithms are efficient on various problem classes, MDP problems and beyond, continue to be filled.

### Acknowledgments

This work was supported in part by NSF grant IIS-9523649 and was performed in large part during the PhD work of the author at the University of Washington. The author is indebted to his advisors Richard Anderson and Steve Hanks for their guidance and support throughout this research. Thanks to Ali Dasdan for valuable discussions and comments on an earlier version of the paper. Many thanks to Russ Greiner and the anonymous referees for their suggestions in improving the presentation.